\documentclass[11pt]{article}

\usepackage[final]{acl}

\usepackage{times}
\usepackage{latexsym}
\usepackage{cleveref}

\usepackage[T1]{fontenc}

\usepackage[utf8]{inputenc}

\usepackage{microtype}

\usepackage{inconsolata} 

\usepackage{graphicx}
\usepackage[table,dvipsnames]{xcolor}   
\usepackage{booktabs}        
\usepackage{multirow}        
\usepackage{graphicx}
\usepackage{enumitem}
\usepackage{placeins}
\usepackage{xurl}
\usepackage{hyperref}
\usepackage{amssymb}
\usepackage[utf8]{inputenc}
\usepackage{booktabs}
\usepackage{pifont}
\usepackage{amssymb}
\usepackage{geometry}
\usepackage{array}
\usepackage{colortbl}
\usepackage{tcolorbox}

\newcommand{\cmark}{\ding{51}}
\newcommand{\xmark}{\ding{55}}

\title{A Shaky Voice Is Not Always a Dodge: Benchmarking Textual and Vocal Evasion Detection in Earnings Calls}

\author{
  Mirae Kim\thanks{\hspace{1mm}Equal contribution.}$^{1}$,
  Seonghun Jeong\footnotemark[1]$^{1,2}$,
  Youngjun Kwak\thanks{\hspace{1mm}Corresponding author.}$^1$ \\
  $^1$Financial Tech Lab, KakaoBank Corp. $^2$Yonsei University\\
  \texttt{\{melissa.kim, bentley.j, vivaan.yjkwak\}@lab.kakaobank.com}
}

\begin{document}
\maketitle
\begin{abstract}
Existing approaches to evasion detection in earnings calls focus on textual transcripts, treating evasion as a single-dimensional phenomenon. We argue that evasion in spoken communication is inherently multidimensional: beyond \emph{what} executives say, \emph{how} they say it carries independent and complementary information. To study these dimensions jointly, we introduce \textbf{DualEvasion}, a benchmark for evasion detection across text and audio in earnings call Q\&A. The benchmark contains 505 annotated question-answer pairs from 60 earnings calls, each with two independent labels: textual evasion (direct vs.\ evasive) and vocal cues operationalized as speaker confidence (confident vs.\ unconfident). Our experiments show that state-of-the-art multimodal models struggle to detect vocal confidence, particularly on unconfident responses. Our analysis suggests these models interpret acoustic cues in isolation rather than relative to each speaker's baseline. Providing speaker-level references yields modest improvements, but a substantial gap with human performance remains.
\end{abstract}

\section{Introduction}
Earnings calls offer a rich source of managerial communication, combining prepared remarks with interactive Q\&A. Prior work has examined diverse signals from these calls, including textual transcripts, acoustic features, and market time-series data, to predict financial outcomes such as stock volatility and return direction~\cite{qin-yang-2019-say, 10.1145/3340531.3412879, sawhney-etal-2020-voltage, yang2022numhtml, 10.1145/3677052.3698689, yu-etal-2025-company}. Q\&A sessions are particularly informative because they are less scripted than prepared remarks and require executives to respond to analysts' questions in real time. This interactive setting has motivated research on evasive behavior in executive responses, showing that evasiveness and semantic deflection can provide early signals of future performance issues and negative market reactions~\cite{10.1287/isre.2022.0415, nuaimi-etal-2025-detecting, hynes2026language}.

\begin{figure}
    \centering
    \includegraphics[width=\linewidth]{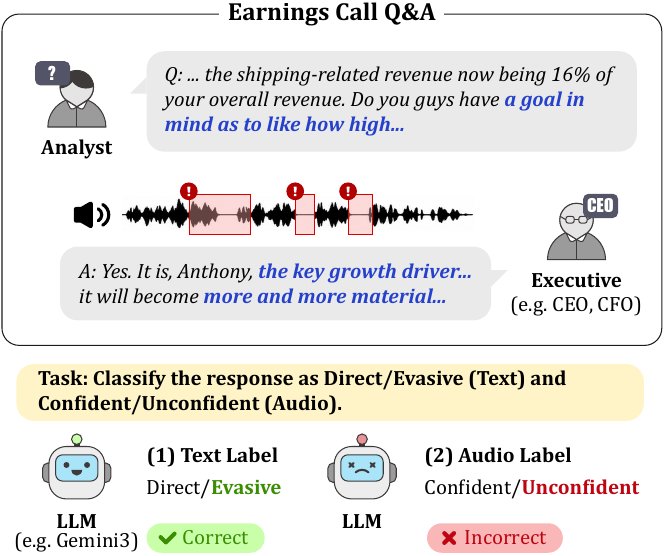}
    \caption{The challenge of \textbf{DualEvasion}: an LLM detects textual evasion but misses vocal unconfidence in the same response.}
    \label{fig:overview}
\end{figure}

However, existing work treats evasion as a purely textual phenomenon, focusing on \emph{what} executives say while overlooking \emph{how} they say it (\Cref{tab:dataset_comparison}). We argue that these two channels carry independent information: an executive may give a textually complete answer yet sound hesitant, or skillfully deflect a question while sounding fully composed. In our data, 32.1\% of instances show cross-modal disagreement---textually direct yet vocally unconfident, or vice versa (\Cref{tab:label_distribution})---indicating that text alone misses an entire dimension of evasive communication. In an exploratory analysis, this missed dimension carries market-relevant signal beyond the transcript (\Cref{sec:market_relevance}).

To capture the vocal dimension, we operationalize it through speaker confidence (confident vs.\ unconfident), grounded in observable prosodic cues that listeners can reliably perceive~\cite{smith1993course, inproceedings, jiang2017sound, goupil2021listeners}. Confidence serves as a concrete, annotatable entry point into the vocal dimension of evasive communication, though it does not exhaust it.  A key challenge is speaker dependence: what sounds hesitant for one executive may reflect another's habitual style, making speaker-aware evaluation essential~\cite{gat2022speaker, van2023modelling}.

We introduce \textbf{DualEvasion}, the first benchmark that provides independent textual and vocal labels for evasion-related behavior in earnings call Q\&A, containing 505 annotated question-answer pairs from 60 calls. Our main contributions are:
 
\begin{itemize} 
\item \textbf{A dual-label evasion benchmark.} We construct the first benchmark for evasion-related behavior in earnings calls with two independent annotation dimensions: textual evasion and vocal confidence. The two dimensions are only weakly correlated, supporting the need to study them jointly. 
\item \textbf{Speaker calibration as a key challenge.} Frontier models struggle with vocal confidence detection (\Cref{fig:overview}), relying on absolute acoustic cues without accounting for each speaker's baseline. Speaker-level calibration yields only modest improvements, revealing a substantial gap with human performance. 
\end{itemize}

\section{Related Work}

\subsection{Textual Evasion}
Prior work has proposed various taxonomies of evasive responses. \citet{rasiah2010framework} categorizes responses as direct, intermediate, or fully evasive, while \citet{bavelas1990equivocal} and \citet{bull1998equivocation} identify finer-grained strategies such as omission, vagueness, and agenda shifting. More recent work has framed evasion detection as a classification task with varying label schemes, in political discourse~\cite{thomas-etal-2024-never, sage2026kclarity} and in earnings calls~\cite{gow2021non, nuaimi-etal-2025-detecting}; however, these studies rely exclusively on textual features. In earnings calls, where analyst questions are goal-directed and executives are expected to provide firm-specific information, we adopt a binary formulation---direct versus evasive---informed by these finer-grained taxonomies but simplified for annotation reliability.

\subsection{Vocal Cues and Confidence}
While textual evasion has been studied across several domains, vocal signals in financial communication remain largely unexplored. Research 
on vocal confidence, however, provides a useful foundation. \citet{smith1993course} show that filled pauses and response delays are associated with lower speaker confidence, while \citet{inproceedings} and \citet{goupil2021listeners} demonstrate that listeners can reliably perceive speaker uncertainty from prosodic cues. \citet{pon2010recognizing} further classify spoken utterances as certain or uncertain using vocal features in dialogue 
settings. These studies establish that confidence-related vocal cues are both perceptually salient and systematically measurable, making speaker confidence a natural operationalization for studying the vocal dimension of earnings call communication.

\begin{table}[t]
\centering
\resizebox{\columnwidth}{!}{%
\renewcommand{\arraystretch}{1.6}
\setlength{\tabcolsep}{8pt}
\begin{tabular}{@{}lcccc@{}}
\toprule
& \textbf{SubjECTive-QA} & \textbf{Nuaimi} & \textbf{EvasionBench} & \textbf{DualEvasion} \\
\midrule
Period     & '07--'21 & '19--'22 & Multi-yr & \textbf{'23--'25} \\
Audio      & \xmark   & \xmark   & \xmark   & \textbf{\cmark} \\
Evasion    & $\triangle$ & \cmark & \cmark  & \textbf{\cmark} \\
Label      & Text     & Text     & Text     & \textbf{Text+Vocal} \\
Annotation & Human    & H+LLM    & LLM      & \textbf{Domain Expert} \\
\bottomrule
\end{tabular}%
}
\caption{Comparison with earnings call Q\&A datasets~\cite{10.5555/3737916.3739810,nuaimi-etal-2025-detecting,ma2026evasionbenchlargescalebenchmarkdetecting}. Existing datasets label evasion from text only; ours additionally captures vocal cues annotated by finance professionals. $\triangle$\,=\,partial.}
\label{tab:dataset_comparison}
\end{table}

\begin{table*}
  \centering
  \setlength{\tabcolsep}{4pt}
  \renewcommand{\arraystretch}{0.9}
  \footnotesize
  \begin{tabular}{l|ccc|ccc|cc}
    \toprule
    \multirow{2}{*}{\textbf{Model}}
      & \multicolumn{3}{c|}{\textbf{Direct/Confident}}
      & \multicolumn{3}{c|}{\textbf{Evasive/Unconfident}}
      & \multicolumn{2}{c}{\textbf{Overall}} \\
    \cmidrule(lr){2-4} \cmidrule(lr){5-7} \cmidrule(lr){8-9}
      & Precision & Recall & F1 & Precision & Recall & F1 & Macro F1 & Accuracy \\
    \midrule
    \rowcolor{green!8}
    \multicolumn{9}{c}{\textit{Textual Evasion: Direct vs. Evasive}} \\
    Qwen2-7B-Instruct       & 79.3          & \underline{91.6} & 85.0          & 60.8          & 35.3          & 44.7          & 64.8          & 76.4          \\
    Llama-3.1-8B-IT         & 84.9          & 80.8          & 82.8          & 53.9          & 61.0          & 57.2          & 70.0          & 75.4          \\
    Gemini-2.5-Flash        & \underline{93.2} & 74.8          & 83.0          & 55.5          & \underline{85.3} & 67.2          & 75.1          & 77.6          \\
    Gemini-3-Flash          & 88.7          & \textbf{97.6} & \textbf{92.9} & \textbf{90.9} & 66.2          & \underline{76.6} & \underline{84.7} & \underline{89.1} \\
    GPT-5                   & \textbf{95.9} & 89.4          & \underline{92.6} & \underline{75.8} & \textbf{89.7} & \textbf{82.2} & \textbf{87.4} & \textbf{89.5} \\
    \midrule
    \rowcolor{red!8}
    \multicolumn{9}{c}{\textit{Vocal Cues: Confident vs. Unconfident}} \\
    Qwen2-Audio-7B-Instruct & 88.0          & 11.1           & 19.7           & 22.4          & \underline{94.4} & 36.2          & 28.0          & 28.9          \\
    Audio-Flamingo-3        & 80.4          & 66.2          & 72.7          & 24.7          & 40.7          & 30.8          & 51.7          & 60.8          \\
    Gemini-2.5-Flash-lite   & 80.6          & 66.0          & 72.6          & 25.0          & 41.7          & 31.2          & 51.9          & 60.8          \\
    Gemini-2.5-Flash        & 83.7          & \underline{72.0} & \underline{77.4} & \underline{32.0}          & 48.4          & 38.5          & \underline{57.9} & \underline{66.9} \\
    Gemini-3-Flash          & 84.0          & 67.8          & 75.0          & 30.7 & 52.5          & \underline{38.7} & 56.8 & 64.5          \\
    Gemini-3.5-Flash        & \underline{88.6} & 33.2          & 48.4          & 25.6          & 84.3          & \textbf{39.2} & 43.8          & 44.2          \\
    GPT-Audio-mini          & \textbf{89.0} & 10.2          & 18.4          & 22.4          & \textbf{95.4} & 36.3          & 27.3          & 28.4          \\
    GPT-Audio               & 83.5          & \textbf{74.6} & \textbf{78.8} & \textbf{32.8} & 45.7          & 38.2          & \textbf{58.5}          & \textbf{68.4} \\
    \bottomrule
  \end{tabular}
  \caption{Zero-shot performance on textual evasion and vocal confidence detection. Results for commercial audio models are averaged over three runs, while all other results are from a single run. \textbf{Bold} indicates best, \underline{underline} indicates second best per column within each section.}
  \label{tab:main_results}
\end{table*}

\definecolor{disagree}{gray}{0.92}
\begin{table}[t]
\centering
\small
\begin{tabular}{@{}llrr@{}}
\toprule
Text & Vocal & Count & \% \\
\midrule
Direct & Confident & 302 & 59.8 \\
\rowcolor{disagree} Direct & Unconfident & 67 & 13.3 \\
\rowcolor{disagree} Evasive & Confident & 95 & 18.8 \\
Evasive & Unconfident & 41 & 8.1 \\
\midrule
\multicolumn{2}{@{}l}{\textbf{Cross-modal disagreement}} & \textbf{162} & \textbf{32.1} \\
\bottomrule
\end{tabular}
\caption{Joint distribution of textual evasion and vocal confidence labels.
Shaded rows indicate cross-modal disagreement.}
\label{tab:label_distribution}
\end{table}

\section{DualEvasion}
\subsection{Data Collection}
We construct \textbf{DualEvasion} from earnings call recordings obtained via the EarningsCall API\footnote{\href{https://earningscall.biz}{\texttt{earningscall.biz}}} under a commercial license. We target 300 tickers spanning 2023--2025, selected using FinanceDataReader\footnote{\href{https://github.com/FinanceData/FinanceDataReader}{\texttt{github.com/FinanceData/FinanceDataReader}}} to ensure diversity across stock exchanges, market capitalization tiers, and industry sectors. To segment each call into question-answer pairs with speaker attribution, we run WhisperX~\cite{bain2022whisperx} on the raw audio for sentence-level timestamps and align the result with transcripts from DefeatBeta\footnote{\href{https://github.com/defeat-beta/defeatbeta-api}{\texttt{github.com/defeat-beta/defeatbeta-api}}}, which provide speaker identity metadata.
Starting from 3,592 calls with available audio, we apply three filtering criteria: (1) calls in which all analyst questions are answered by a single executive, to avoid ambiguity in vocal cue assessment (110 calls); (2) at least four Q\&A pairs per call, to ensure sufficient per-speaker reference material (96 calls); and (3) all responses under 300 seconds, to fit within model input constraints (93 calls). From these, we select 60 calls across 49 unique tickers. Detailed statistics on the dataset composition are provided in \Cref{app:dataset_statistics}.

\subsection{Annotation and Validation}
The final benchmark contains 505 question-answer pairs from the 60 selected calls, each annotated along two dimensions: textual evasion and vocal confidence. Two financial domain experts independently annotate all pairs, followed by a separate reliability check on a subset. Annotation details are in \Cref{app:annotation}.
 
\paragraph{Textual Evasion Labels.}
Each response is labeled as either \textit{direct} or \textit{evasive} based on whether the executive substantively addresses the analyst's question, with responses that avoid, deflect, or only partially address the core issue classified as evasive (Cohen's $\kappa$\footnote{Cohen's $\kappa$ \citep{cohen1960coefficient} measures inter-rater agreement beyond chance, ranging from $-1$ (systematic disagreement) to $1$ (perfect agreement).} $= 0.866$, 94.7\% agreement).
 
\paragraph{Vocal Confidence Labels.}
Vocal confidence labels are assigned in a separate pass: annotators listen to audio responses only, without reference to the textual labels or transcript content. Each response is labeled as either \textit{confident} or \textit{unconfident} based on prosodic cues such as filled pauses, hesitation, and intonation patterns. To account for speaker-specific vocal characteristics, annotators first listen to the executive's other responses within the same call to establish a per-speaker baseline before making judgments (Cohen's $\kappa$ = 0.774, 87.5\% agreement).
 
\paragraph{Validation.}
To check whether agreement generalizes beyond the two primary annotators, five financial domain experts---including the original two, using their pre-adjudication labels---independently label a random subset of 52 responses for vocal confidence, yielding Fleiss' $\kappa$\footnote{Fleiss' $\kappa$~\citep{article} extends Cohen's $\kappa$ to three or more raters.} of 0.713 (substantial agreement; pairwise $\kappa$ in \Cref{tab:pairwise}). These validation labels serve only this reliability check and do not affect the final labels, which are produced by the two experts through adjudication. The resulting label distribution is shown in \Cref{tab:label_distribution}; the two dimensions are only weakly correlated, with 32.1\% of instances showing cross-modal disagreement.

\begin{table}[t]
\centering
\small
\begin{tabular}{lcc}
\toprule
 & \textbf{Text} & \textbf{Vocal} \\
\midrule
LLM--Annotator          & 0.813 & 0.225 \\
Annotator--Annotator    & 0.866 & 0.774 \\
\bottomrule
\end{tabular}
\caption{Cohen's $\kappa$ agreement with human annotators, averaged over two annotators, and human--human agreement. GPT-5 is used for text and Gemini-3-Flash for vocal confidence.}
\label{tab:agreement}
\end{table}

\section{Experiment}
We evaluate models across both dimensions in a zero-shot setting, using the human annotations as ground truth. For textual evasion, we test open-source and commercial language models on transcript text. For vocal confidence, we test audio-capable models on response audio. Main results are reported in \Cref{tab:main_results}, with all prompts in \Cref{app:prompt}.

\subsection{A Modality Gap in Model Performance}
\Cref{tab:main_results} reveals a clear modality gap. For textual evasion, frontier models perform well: GPT-5 and Gemini-3-Flash reach 87.4 and 84.7 macro F1, respectively. Vocal confidence detection, however, proves far more challenging---the best audio model achieves only 58.5 macro F1. As \Cref{tab:agreement} shows, GPT-5's agreement with annotators on textual evasion approaches human--human levels, whereas Gemini-3-Flash's agreement on vocal confidence remains substantially lower.

The major bottleneck lies in the unconfident class, where F1 ranges from 30.8 to 39.2---far below the confident class (up to $\sim$79 F1). This weakness cannot be explained by class imbalance alone: the strongest models exceed the 44.0 macro-F1 all-confident majority baseline (e.g., GPT-Audio at 58.5), while weaker models fail in the opposite direction, overpredicting the unconfident class with recall as high as 95.4. Neither pattern is consistent with a simple majority-class bias. Threshold tuning and supervised fine-tuning likewise fail to close the gap (\Cref{app:finetuning}). These results point to a persistent difficulty in detecting unconfidence, which we investigate further in \Cref{sec:speaker_calibration}.

\begin{figure}[t]
    \centering
    \includegraphics[width=\linewidth]{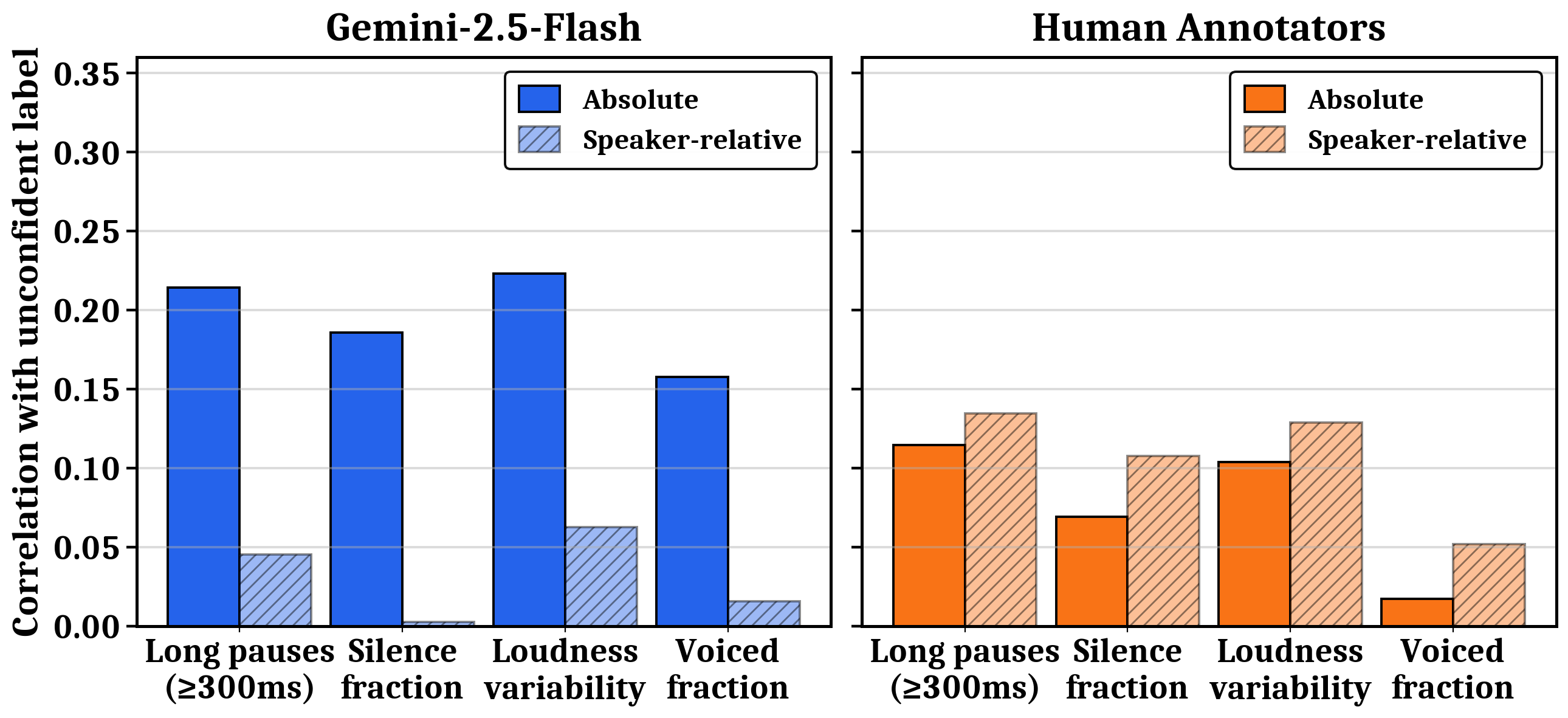}
    \caption{Correlation between acoustic features and unconfident predictions for Gemini-2.5-Flash, the model with the largest gap, measured using absolute values and speaker-relative deviations (response value minus the speaker's mean).}
    \label{fig:acoustic}
\end{figure}

\begin{table}[t]
\centering
\small
\setlength{\tabcolsep}{5pt}
\begin{tabular}{lccc}
\toprule
 & $|r_{\mathrm{raw}}|$ & $|r_{\mathrm{cent}}|$ & \textbf{Gap} ($p$) \\
\midrule
\rowcolor{RoyalBlue!12}
\multicolumn{4}{l}{\textit{Models}} \\
Gemini-2.5-Flash      & .195 & .032 & \textbf{.164 (.005)} \\
Gemini-3-Flash        & .179 & .045 & \textbf{.135 (.003)} \\
GPT-Audio             & .142 & .030 & \textbf{.113 (.025)} \\
Gemini-2.5-Flash-lite & .093 & .013 & .080 (.234) \\
Audio-Flamingo-3      & .040 & .015 & .025 (.522) \\
\midrule
\rowcolor{Orange!18}
\multicolumn{4}{l}{\textit{Human annotators}} \\
A1        & .074 & .099 & $-.025$ (.510) \\
A2        & .079 & .112 & $-.034$ (.398) \\
Consensus & .088 & .104 & $-.016$ (.654) \\
\bottomrule
\end{tabular}
\caption{Mean absolute point-biserial correlation between acoustic features and the unconfident label, before ($|r_{\mathrm{raw}}|$) and after ($|r_{\mathrm{cent}}|$) speaker-centering. Gap $= |r_{\mathrm{raw}}| - |r_{\mathrm{cent}}|$; $p$-values from a cluster bootstrap over the 60 calls. \textbf{Bold} indicates gaps significant at $p<.05$.}
\label{tab:centering}
\end{table}

\begin{table}[t]
  \centering
  \small
  \setlength{\tabcolsep}{6pt}
  \begin{tabular}{l|ccc}
    \toprule
    \textbf{Model} & \textbf{Baseline} & \textbf{+Speaker Norm.} & \textbf{$\Delta$F1} \\
    \midrule
    Gemini-2.5-Flash & 38.5 & 43.2 & \textbf{+4.7} \\
    Gemini-3-Flash   & 38.7 & 41.6 & \textbf{+2.9} \\
    GPT-Audio        & 38.2 & 39.2 & \textbf{+1.0} \\
    \bottomrule
  \end{tabular}
  \caption{Effect of speaker-level normalization on unconfident F1. The model infers the speaker's vocal baseline before classifying the target response.}
  \label{tab:speaker_norm}
\end{table}

\subsection{Speaker Calibration}
\label{sec:speaker_calibration}
\paragraph{Acoustic analysis.}
One explanation for this difficulty is that models judge confidence from raw acoustic values rather than relative to each speaker's baseline. To test this, we measure the point-biserial correlation between four acoustic features and the unconfident label under two conditions: absolute feature values and speaker-relative deviations (each response minus the speaker's mean). We report the mean absolute correlation across features before ($|r_{\mathrm{raw}}|$) and after ($|r_{\mathrm{cent}}|$) speaker-centering, with significance estimated by a cluster bootstrap over the 60 calls (\Cref{tab:centering}).

Across the three strongest models, speaker-centering sharply reduces the correlation---for Gemini-2.5-Flash, from $.195$ to $.032$ ($p<.01$)---indicating reliance on raw acoustic values rather than each speaker's baseline (\Cref{fig:acoustic}). Human annotators show no such drop; their correlations remain stable or increase slightly, consistent with speaker-aware judgment. Weaker models show smaller gaps, but their raw correlations are already near zero, so little remains to remove. Feature definitions and extraction details are in \Cref{app:acoustic}.

\paragraph{Calibration experiments.}
Given this apparent reliance on absolute cues, we test whether providing explicit speaker context helps. In our main evaluation, models receive only a single response audio, with no access to other utterances from the same speaker. We first provide same-speaker reference audio as few-shot examples, but this yields no consistent improvement (results in \Cref{app:fewshot}). We then prompt models to first infer the speaker's habitual vocal baseline (such as pitch, tempo, and prosody) from three same-speaker reference utterances---selected sequentially and excluding the target---before judging whether the target response deviates from it (prompt template in \Cref{tab:prompt-speaker-norm}). As shown in \Cref{tab:speaker_norm}, this modestly improves unconfident F1 (e.g., +4.7 for Gemini-2.5-Flash), suggesting that speaker-level calibration is a promising direction, though the gap with human performance remains large.

\subsection{Market Relevance}
\label{sec:market_relevance}
\begin{figure}[t]
    \centering
    \includegraphics[width=\linewidth]{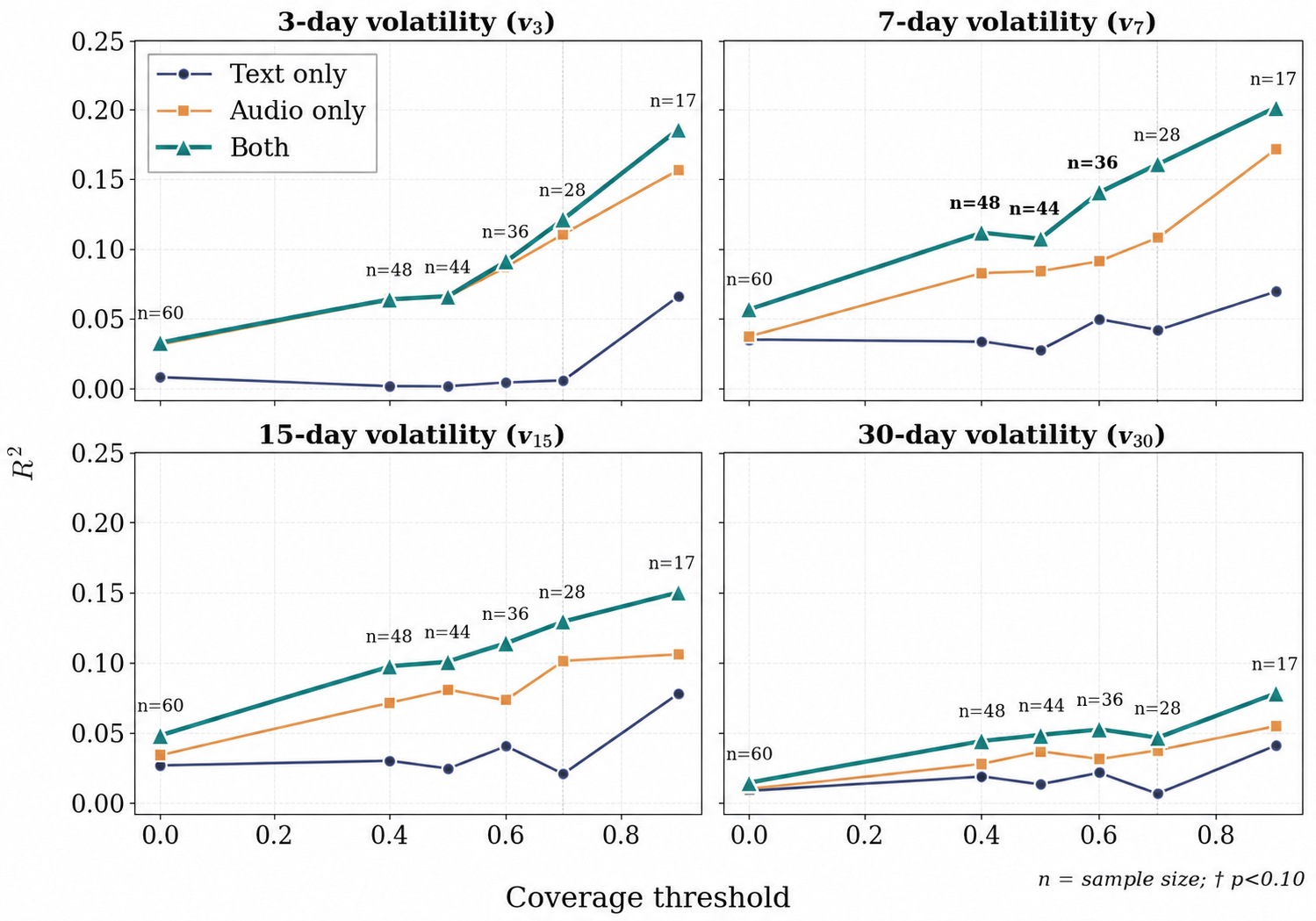}
    \caption{$R^2$ of text-only, audio-only, and combined regression models for post-earnings stock volatility across Q\&A coverage thresholds. Coverage is the fraction of each call's Q\&A pairs included in \textbf{DualEvasion}; at each threshold, only calls with coverage at or above that level are retained. Annotations indicate the number of retained calls ($n$); $\dagger$: $p<0.10$.}
    \label{fig_stockanalysis}
\end{figure}

To examine whether our labels carry market-relevant information, we regress call-level evasive and unconfident ratios against post-earnings stock volatility over 3- to 30-day windows (\Cref{fig_stockanalysis}). Because \textbf{DualEvasion} contains only a subset of Q\&A pairs for some calls, we repeat the analysis while progressively restricting the sample to calls with more complete Q\&A coverage. Across these coverage settings, audio-based unconfidence generally explains more variance than textual evasion, and combining both yields the highest $R^2$, suggesting that the two dimensions capture complementary signals. This association is strongest over shorter horizons ($v_3$, $v_7$) and weakens by $v_{30}$. Given the limited sample of 60 calls, we treat these results as exploratory rather than causal; regression and volatility details are provided in \Cref{app:volatility}.

\section{Conclusion}
We introduced \textbf{DualEvasion}, a benchmark for evasion detection across text and audio in earnings call Q\&A, containing 505 annotated question-answer pairs with independent textual evasion and vocal confidence labels. Our experiments reveal that models approach human-level performance on textual evasion but fall far short on vocal confidence detection. Acoustic analysis shows that models rely on absolute vocal features rather than speaker-relative deviations, conflating habitual speaker traits with uncertainty cues. Speaker-level calibration yields modest improvements but does not close the gap, suggesting that speaker-aware audio understanding remains an open challenge for current models.

\section{Limitations}
Our benchmark contains 505 question-answer pairs from 60 calls. The dual-annotation design---requiring independent textual and vocal labeling, with vocal annotation further requiring per-speaker calibration across all responses in the same call---makes annotation substantially more time-intensive than standard labeling tasks, limiting the scale we could 
achieve. Our vocal dimension is operationalized through speaker confidence, which captures one aspect of vocal cues but does not exhaust them. Finally, our stock volatility analysis is based on a small sample and should be interpreted as exploratory.

\bibliography{custom}

\appendix

\FloatBarrier
\section{Dataset Statistics}
\label{app:dataset_statistics}
We summarize the dataset composition in \Cref{tab:dataset_composition} and response-level statistics in \Cref{tab:response_stats}. The benchmark spans 11 sectors and three market groups, with response durations ranging from under 1 second to nearly 5 minutes.

\begin{table}[htbp]
\centering
\small
\begin{tabular}{@{}lr@{}}
\toprule
\multicolumn{2}{l}{\textbf{Composition}} \\
\midrule
Earnings calls & 60 \\
Q\&A pairs & 505 \\
Unique tickers & 49 \\
Call period & 2023--2025 \\
Market cap range & \$56M--\$189B \\
\midrule
\multicolumn{2}{l}{\textbf{Market Group}} \\
\midrule
NYSE & 27 (45\%) \\
NASDAQ & 17 (28\%) \\
S\&P 500 & 16 (27\%) \\
\midrule
\multicolumn{2}{l}{\textbf{Sector}} \\
\midrule
Technology & 13 \\
Industrials & 10 \\
Financial Services & 8 \\
Energy & 5 \\
Consumer Cyclical & 5 \\
Healthcare & 4 \\
Basic Materials & 4 \\
Communication Svcs. & 4 \\
Real Estate & 3 \\
Consumer Defensive & 2 \\
Utilities & 2 \\
\bottomrule
\end{tabular}
\caption{Dataset composition.}
\label{tab:dataset_composition}
\end{table}

\begin{table}[htbp]
\centering
\small
\begin{tabular}{@{}lrrr@{}}
\toprule
 & Med. & Min & Max \\
\midrule
Response duration (sec) & 66.9 & 0.2 & 296.9 \\
Response length (tokens) & 186 & 2 & 918 \\
Q\&A pairs per call & 10 & 4 & 10 \\
\midrule
\multicolumn{3}{@{}l}{Total audio duration} & 10.9 hrs \\
\bottomrule
\end{tabular}
\caption{Response duration and length statistics.}
\label{tab:response_stats}
\end{table}

\FloatBarrier
\section{Annotation Details}
\label{app:annotation}
\subsection{Annotation Guidelines}
The following guidelines were provided to all annotators for both textual evasion and vocal confidence labeling (\Cref{tab:guideline-text,tab:guideline-audio}).

\subsection{Annotation Interface}
\begin{figure}
    \centering
    \includegraphics[width=\linewidth]{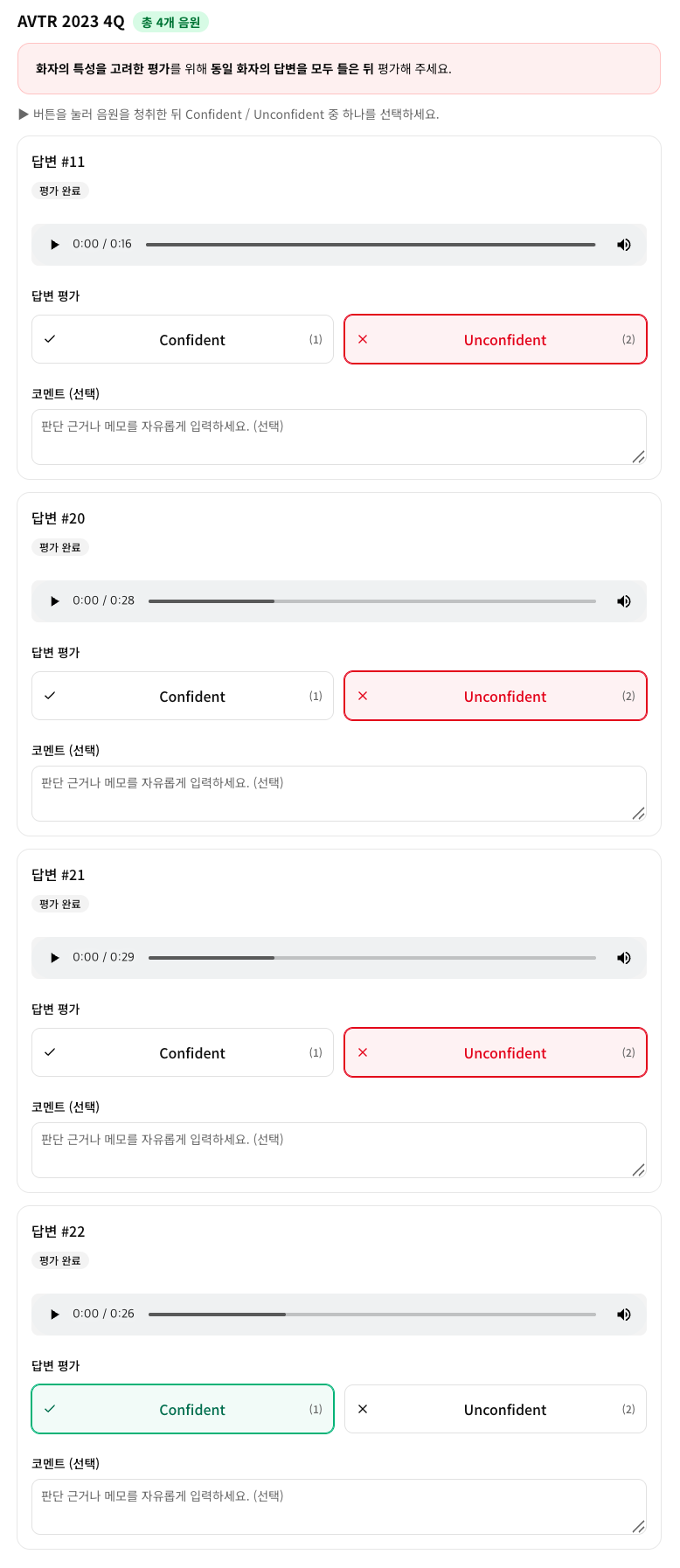}
    \caption{Annotation interface for vocal confidence.}
    \label{fig:annotation-interface}
\end{figure}

\Cref{fig:annotation-interface} shows the annotation interface used for vocal confidence labeling. Annotators listen to each response audio and select 
one of the two labels. A comment field is provided for ambiguous cases.

\subsection{Inter-Annotator Agreement}
\label{app:agreement}
\Cref{tab:pairwise} reports pairwise Cohen's $\kappa$ among the five validation annotators for vocal confidence. We conducted this additional validation only for vocal confidence, as textual evasion already achieved 94.7\% raw agreement in the main annotation. The comment field was optional; nine comments were provided in total, all for unconfident judgments, citing cues such as a large sigh, slower-than-usual speech, or pauses within the utterance. For comparison, \Cref{tab:pairwise_model} reports pairwise agreement among the three strongest audio models, which is markedly lower than human inter-annotator agreement, further illustrating the difficulty of vocal confidence detection.

\begin{table}[htbp]
  \centering
  \small
  \begin{tabular}{@{}lccccc@{}}
    \toprule
     & \textbf{A1} & \textbf{A2} & \textbf{A3} & \textbf{A4} & \textbf{A5} \\
    \midrule
    \textbf{A1} & -- & .915 & .706 & .651 & .917 \\
    \textbf{A2} &    & --   & .623 & .651 & .834 \\
    \textbf{A3} &    &      & --   & .442 & .713 \\
    \textbf{A4} &    &      &      & --   & .662 \\
    \textbf{A5} &    &      &      &      & --   \\
    \bottomrule
  \end{tabular}
  \caption{Pairwise Cohen's $\kappa$ for vocal confidence labels on the validation subset ($N$ = 52).}
  \label{tab:pairwise}
\end{table}

\begin{table}[htbp]
  \centering
  \small
  \begin{tabular}{@{}lccccc@{}}
    \toprule
     & \textbf{Gem-2.5} & \textbf{Gem-3.0} & \textbf{GPT-Aud} \\
    \midrule
    \textbf{Gem-2.5} & -- & .483 & .309 \\
    \textbf{Gem-3.0}   &    & --   & .407 \\
    \textbf{GPT-Aud}        &    &      & --   \\
    \bottomrule
  \end{tabular}
  \caption{Inter-model agreement on vocal confidence labels measured by pairwise Cohen’s $\kappa$ ($N$ = 52). Gem-2.5 denotes Gemini-2.5-Flash, Gem-3.0 denotes Gemini-3-Flash, and GPT-Aud denotes GPT-Audio.}
  \label{tab:pairwise_model}
\end{table}

\begin{table*}[t]
\centering
\small
\fbox{\begin{minipage}{0.95\textwidth}
\textbf{Textual Evasion Annotation Guidelines}

\vspace{6pt}
Evaluate whether the executive substantively addresses the analyst's question. Assess only the textual content, not vocal delivery.

\vspace{6pt}
\textbf{Direct}: The response substantively addresses the core question with a relevant and specific explanation, even if no exact figures are provided.

\vspace{6pt}
\textbf{Evasive}: The response fails to address the core question. Label as evasive if the response exhibits any of the following patterns:

\vspace{4pt}
\hspace*{1em}\textit{General forms}:
\begin{itemize}[nosep,leftmargin=2em]
\item Omission of key information
\item Vague or non-specific language
\item Non-sequitur (unrelated response)
\item Restatement of the question or prior remarks
\end{itemize}

\vspace{4pt}
\hspace*{1em}\textit{Specific strategies}:
\begin{itemize}[nosep,leftmargin=2em]
\item Avoidance or deflection
\item Acknowledging the question without answering it
\item Explicit refusal to answer
\item Shifting to a different topic
\item Claiming ignorance or lack of information
\item Providing only a partial answer or selectively disclosing information
\item Interpreting the question too literally
\item Repeating previously stated material
\item Challenging the premise of the question
\item Questioning the question itself or the questioner
\item Attributing issues to external factors
\end{itemize}

\vspace{6pt}
\textbf{Important:}
\begin{itemize}[nosep,leftmargin=1.5em]
\item Explicitly declining to answer (e.g., ``we can't disclose,'' ``I don't have that number'') is \textit{evasive}.
\item If uncertain, choose the closer label and leave a comment.
\end{itemize}
\end{minipage}}
\caption{Annotation guidelines for textual evasion.}
\label{tab:guideline-text}
\end{table*}

\begin{table*}
\centering
\small
\fbox{\begin{minipage}{0.95\textwidth}
\textbf{Vocal Confidence Annotation Guidelines}

\vspace{6pt}
Evaluate the executive's vocal attitude. The content of the answer is \textbf{not} the subject of evaluation. Assess only the speaker's confidence level as expressed through their voice.

\vspace{6pt}
\textbf{Confident}:
\begin{itemize}[nosep,leftmargin=1.5em]
\item Stable and clear voice
\item Consistent speech rate and rhythm
\item Minimal fillers or stuttering
\item Fluent and natural delivery
\item Calm and controlled overall tone
\end{itemize}

\vspace{6pt}
\textbf{Unconfident}:
\begin{itemize}[nosep,leftmargin=1.5em]
\item Unstable or trembling voice
\item Frequent or prolonged pauses
\item Frequent fillers (um, uh)
\item Irregular speech rate or stuttering
\item Tense overall tone
\end{itemize}

\vspace{6pt}
\textbf{Important:}

\vspace{4pt}
\textbf{1. Consider speaker characteristics.}\\
Listen to all responses from the same speaker before rating.
\begin{itemize}[nosep,leftmargin=1.5em]
\item If a speaker habitually speaks slowly, do not label them unconfident solely for slow tempo.
\item If a speaker frequently uses fillers (e.g., ``um''), do not label them unconfident for that reason alone.
\end{itemize}

\vspace{4pt}
\textbf{2. Evaluate voice, not content.}\\
Focus on \textit{how} the executive speaks, not \textit{what} they say.
\begin{itemize}[nosep,leftmargin=1.5em]
\item A negative message delivered in a stable voice is \textit{confident}.
\item A positive message delivered with trembling or stuttering is \textit{unconfident}.
\end{itemize}

\vspace{4pt}
\textbf{3. Always choose one label.}\\
If uncertain, choose the closer option and leave a comment.
\end{minipage}}
\caption{Annotation guidelines for vocal confidence.}
\label{tab:guideline-audio}
\end{table*}

\FloatBarrier
\section{Additional Baselines and Fine-Tuning}
\label{app:finetuning}
\subsection{Baseline and Threshold Analysis}
A majority (all-confident) baseline yields only 44.0 macro F1 (confident F1 88.0, unconfident F1 0.0; accuracy 78.6). The strongest audio models exceed this baseline, while others fall below it by overpredicting the unconfident class, so the low scores are not a simple majority-class artifact.

We further check whether the decision threshold explains the weak unconfident performance. Commercial models return only discrete labels, so we sweep the decision threshold from 0.05 to 0.95 for the two open models that expose logits. \Cref{tab:threshold} reports, for each model, the threshold maximizing macro F1 and the threshold maximizing unconfident F1. Even at the threshold that maximizes unconfident F1, performance reaches only 36.7 and 35.5, barely above the untuned scores of 36.2 and 30.8. Further prioritizing unconfident detection sharply degrades confident-class performance and drives macro F1 below the 44.0 majority baseline. Threshold adjustment thus shifts the precision--recall trade-off but does not resolve the weak unconfident performance.

\begin{table}[t]
\centering
\small
\setlength{\tabcolsep}{4pt}
\resizebox{\columnwidth}{!}{%
\begin{tabular}{llccc}
\toprule
\textbf{Model} & \textbf{Thr.} & \textbf{Conf. F1} & \textbf{Unconf. F1} & \textbf{Macro F1} \\
\midrule
\multirow{2}{*}{Audio-Flamingo-3}
  & 0.40$^{\mathrm{M}}$ & 82.7 & 30.0 & \textbf{56.4} \\
  & 0.95$^{\mathrm{U}}$ & 2.5 & \textbf{35.5} & 19.0 \\
\midrule
\multirow{2}{*}{Qwen2-Audio}
  & 0.05$^{\mathrm{M}}$ & 65.2 & 34.3 & \textbf{49.7} \\
  & 0.45$^{\mathrm{U}}$ & 24.4 & \textbf{36.7} & 30.5 \\
\bottomrule
\end{tabular}%
}
\caption{Threshold sweep for the two open models that expose logits. $^{\mathrm{M}}$: threshold maximizing macro F1; $^{\mathrm{U}}$: threshold maximizing unconfident F1. \textbf{Bold} marks the maximized metric.}
\label{tab:threshold}
\end{table}

\subsection{Supervised Fine-Tuning}
To test whether the weak vocal confidence performance can be overcome with task-specific supervision, we fine-tune three open speech models on \textbf{DualEvasion}. Given the limited dataset size (505 pairs), we use five-fold cross-validation, ensuring that every instance is evaluated on a held-out fold. \Cref{tab:finetune} summarizes the results.

Fine-tuning improves overall macro F1 for some models, most notably Qwen2-Audio ($+20.6$), but these gains do not translate into better detection of the unconfident class. For Qwen2-Audio, the improvement largely comes from recovering confident-class performance from its near-collapse in the zero-shot setting, while unconfident F1 decreases from 36.2 to 32.4. Audio-Flamingo-3 shows only a modest gain in macro F1 ($+1.8$), with unconfident F1 remaining low at 34.7. VibeVoice-ASR similarly fails on the minority class, reaching only 9.9 unconfident F1 despite a macro F1 of 48.8. Across all fine-tuned models, unconfident F1 remains at or below 34.7, suggesting that task-specific supervision alone does not close the gap in unconfident detection.

\begin{table}[t]
\centering
\footnotesize
\setlength{\tabcolsep}{2.5pt}
\begin{tabular}{ll r@{\;}l r@{\;}l}
\toprule
\textbf{Model} & \textbf{Setting}
& \multicolumn{2}{c}{\textbf{Unconf. F1}}
& \multicolumn{2}{c}{\textbf{Macro F1}} \\
\midrule
\multirow{2}{*}{Qwen2-Audio}
  & zero-shot & 36.2 & {} & 28.0 & {} \\
  & fine-tuned & 32.4 & \textcolor{red}{\scriptsize(-3.8)} & 48.6 & \textcolor{blue}{\scriptsize(+20.6)} \\
\midrule
\multirow{2}{*}{Audio-Flamingo-3}
  & zero-shot & 30.8 & {} & 51.7 & {} \\
  & fine-tuned & 34.7 & \textcolor{blue}{\scriptsize(+3.9)} & 53.5 & \textcolor{blue}{\scriptsize(+1.8)} \\
\midrule
VibeVoice-ASR
  & fine-tuned & 9.9 & {} & 48.8 & {} \\
\bottomrule
\end{tabular}
\caption{Supervised fine-tuning with five-fold cross-validation.
Values in parentheses indicate changes from zero-shot performance.
Fine-tuning improves overall performance but does not consistently improve unconfident detection.}
\label{tab:finetune}
\end{table}

\section{Acoustic Feature Extraction}
\label{app:acoustic}
All features are extracted from 16\,kHz mono audio using \texttt{librosa} and \texttt{parselmouth} (Praat). Frame-level RMS energy (root-mean-square amplitude) is computed with 25\,ms windows and 10\,ms hops. For each feature, we compute both the \textbf{absolute} value and the \textbf{speaker-relative} value (response value minus the speaker's mean across all responses in the same call).

\paragraph{Silence fraction.}
RMS energy is converted to dB and peak-normalized. Frames below $-$35\,dB are marked as silent. Silence fraction is the proportion of silent frames.

\paragraph{Long pauses.}
From the silence mask above, we identify contiguous silent segments $\geq$300\,ms (30 frames). Long pause fraction is the total duration of these segments divided by response duration.

\paragraph{Loudness variability.}
Standard deviation of the peak-normalized RMS energy in dB across all frames.

\paragraph{Voiced fraction.}
Fundamental frequency (F0) is estimated using Praat's autocorrelation method (\texttt{to\_pitch}, pitch floor = 75\,Hz, ceiling = 400\,Hz, time step = 10\,ms). Voiced fraction is the proportion of frames where F0 $>$ 0, i.e., where vocal fold vibration is detected.

\begin{table*}[t]
\centering
\small
\setlength{\tabcolsep}{4pt}
\begin{tabular}{llccccccc}
\toprule
\multirow{2}{*}{\textbf{Model}} & \multirow{2}{*}{\textbf{Methods}} & \multicolumn{3}{c}{\textbf{Confident}} & \multicolumn{3}{c}{\textbf{Unconfident}} & \multirow{2}{*}{\textbf{Macro-F1}} \\
\cmidrule(lr){3-5} \cmidrule(lr){6-8}
 & & Precision & Recall & F1 & Precision & Recall & F1 & \\
\midrule
\multirow{4}{*}{Gemini-2.5-Flash}
 & baseline         & \textbf{83.7} & \cellcolor{gray!20}72.0 & \cellcolor{gray!20}77.4 & \cellcolor{gray!20}32.0 & \textbf{48.4} & 38.5 & 57.9 \\
 & fewshot($n{=}1$) & 83.6 & 77.1 & 80.2 & 34.5 & 44.4 & \textbf{38.9} & \textbf{59.5} \\
 & fewshot($n{=}2$) & \cellcolor{gray!20}81.1 & 85.6 & 83.3 & 34.1 & 26.9 & 30.1 & \cellcolor{gray!20}56.7 \\
 & fewshot($n{=}3$) & \cellcolor{gray!20}81.1 & \textbf{91.9} & \textbf{86.2} & \textbf{41.8} & \cellcolor{gray!20}21.3 & \cellcolor{gray!20}28.2 & 57.2 \\
\midrule
\multirow{4}{*}{Gemini-3-Flash}
 & baseline         & 84.0 & \cellcolor{gray!20}67.8 & \cellcolor{gray!20}75.0 & \cellcolor{gray!20}30.7 & \textbf{52.5} & 38.7 & \cellcolor{gray!20}56.8 \\
 & fewshot($n{=}1$) & \textbf{84.2} & 81.9 & 83.0 & 39.5 & 43.5 & \textbf{41.4} & \textbf{62.2} \\
 & fewshot($n{=}2$) & 82.5 & 88.2 & \textbf{85.3} & \textbf{42.0} & 31.5 & 36.0 & 60.6 \\
 & fewshot($n{=}3$) & \cellcolor{gray!20}82.2 & \textbf{88.4} & 85.2 & 41.0 & \cellcolor{gray!20}29.6 & \cellcolor{gray!20}34.4 & 59.8 \\
\midrule
\multirow{4}{*}{GPT-Audio}
 & baseline         & \textbf{83.5} & 74.6 & \cellcolor{gray!20}78.8 & 32.8 & \textbf{45.7} & \textbf{38.2} & \textbf{58.5} \\
 & fewshot($n{=}1$) & 80.9 & 84.1 & 82.5 & 31.5 & 26.9 & 29.0 & 55.7 \\
 & fewshot($n{=}2$) & 81.1 & \textbf{86.4} & \textbf{83.7} & \textbf{34.1} & 25.9 & 29.5 & 56.6 \\
 & fewshot($n{=}3$) & \cellcolor{gray!20}79.8 & 84.4 & 82.0 & \cellcolor{gray!20}27.1 & \cellcolor{gray!20}21.3 & \cellcolor{gray!20}23.8 & \cellcolor{gray!20}52.9 \\
\bottomrule
\end{tabular}
\caption{Effect of few-shot same-speaker reference audio on vocal confidence detection. \textbf{Bold} indicates the best score and \colorbox{gray!20}{shaded} indicates the lowest score within each model.}
\label{tab:fewshot_results}
\end{table*}

\FloatBarrier
\section{Few-Shot Speaker Calibration}
\label{app:fewshot}
A natural strategy for improving vocal confidence detection is to mimic how human annotators work: before judging a target response, first listen to several other responses from the same speaker to establish a personal baseline. We operationalize this by prepending $n \in \{1, 2, 3\}$ same-speaker reference audio clips from the same earnings call before the target clip and asking the model to judge the target relative to these references. The corresponding prompt is provided in \Cref{tab:prompt-fewshot}.

\Cref{tab:fewshot_results} shows that this approach yields no consistent improvement. While Gemini-2.5-Flash and Gemini-3-Flash see modest gains in macro-F1 at $n{=}1$ (+1.6 and +5.4, respectively), performance on the unconfident class degrades as more reference samples are added: unconfident F1 drops from 38.5 to 28.2 for Gemini-2.5-Flash at $n{=}3$, and from 38.2 to 23.8 for GPT-Audio. Across all models, additional reference audio generally increases confident recall relative to the baseline, while reducing unconfident recall, suggesting that the added context shifts predictions toward the confident class rather than improving sensitivity to speaker-relative deviations.

One possible explanation is that current audio LLMs struggle to form a stable speaker-level baseline from in-context examples. Unlike human annotators, who can integrate multiple utterances into a robust sense of a speaker's typical delivery, models may remain sensitive to the particular reference clips provided. As a result, simply adding more same-speaker examples may not provide a consistent basis for identifying subtle deviations in the target response.

\section{Stock Volatility Analysis}
\label{app:volatility}

\paragraph{Return and volatility computation.}
Daily returns are computed from closing prices as:
\begin{equation}
  r_t = \frac{P_t - P_{t-1}}{P_{t-1}}
\end{equation}
where $P_t$ is the closing price on trading day $t$. Volatility over a $k$-day window is then:
\begin{equation}
  v_k = \log \left( \sqrt{\frac{\sum_{i=1}^{k}(r_i - \bar{r})^2}{k}} \right)
\end{equation}
where $\bar{r}$ is the mean return over the window.
 
\paragraph{Trading Day 0.}
We define Trading Day 0 based on the earnings call timing relative to market hours. For calls released before market open (BMO), the announcement date is Trading Day 0. For calls released after market close (AMC) or during market hours (DMH), the next trading day is Trading Day 0. This convention ensures a full post-call trading session and avoids mixing pre-call and post-call price movements for DMH calls.

\paragraph{Coverage threshold.}
We define Q\&A coverage as the fraction of all Q\&A pairs in a call that are included in \textbf{DualEvasion}. For each coverage threshold, we retain only calls with coverage at or above that level and rerun the regression analysis. For example, a threshold of 0.6 includes only calls for which at least 60\% of the Q\&A pairs are represented in our benchmark. Higher thresholds therefore provide more complete call-level estimates but reduce the number of calls available for analysis (\Cref{fig_stockanalysis}).

\FloatBarrier
\section{Prompt Templates}
\label{app:prompt}
\Cref{tab:prompt-text} and \Cref{tab:prompt-audio} show the prompts used for the main textual evasion and vocal confidence evaluations, respectively. \Cref{tab:prompt-fewshot} and \Cref{tab:prompt-speaker-norm} show the prompts used for the few-shot and speaker-normalized calibration experiments.

\begin{table}[htbp]
\centering
\begin{tcolorbox}[
  width=0.95\columnwidth,
  colback=gray!5,
  colframe=gray!50,
  boxrule=0.6pt,
  arc=3pt,
  left=6pt, right=6pt, top=4pt, bottom=5pt,
  title=\textbf{Textual Evasion Classification},
  coltitle=white,
  colbacktitle=gray!75,
  fonttitle=\footnotesize,
  fontupper=\footnotesize
]
You are a financial analyst evaluating evasiveness in earnings call Q\&A.

\vspace{4pt}
{\color{blue!60}\textbf{Input}}\\
You will receive a Question and an Answer as text. Evaluate semantic content --- specificity, relevance, and completeness of the response.

\vspace{4pt}
{\color{blue!60}\textbf{Classification}}\\
$\bullet$ \texttt{direct}: Addresses the core question with a relevant and substantive explanation, even if no exact figures are provided.\\
$\bullet$ \texttt{evasive}: Fails to address the core question. This includes: shifting to a different topic, omitting key details, giving vague or generic statements, or explicitly refusing to answer.

\vspace{4pt}
{\color{blue!60}\textbf{Output Format}}\\
Classification: \texttt{<direct | evasive>}

\vspace{4pt}
Now classify the following:\\
Question: \texttt{\{question\}}\\
Answer: \texttt{\{answer\}}
\end{tcolorbox}
\caption{Prompt for textual evasion classification.}
\label{tab:prompt-text}
\end{table}


\begin{table}[htbp]
\centering
\begin{tcolorbox}[
  width=0.95\columnwidth,
  colback=gray!5,
  colframe=gray!50,
  boxrule=0.6pt,
  arc=3pt,
  left=6pt, right=6pt, top=4pt, bottom=5pt,
  title=\textbf{Vocal Confidence Classification},
  coltitle=white,
  colbacktitle=gray!75,
  fonttitle=\footnotesize,
  fontupper=\footnotesize
]
You are a financial analyst evaluating vocal confidence in earnings call Q\&A.

\vspace{4pt}
{\color{blue!60}\textbf{Input}}\\
You will receive an audio recording of an executive's response during an earnings call. Evaluate the speaker's vocal confidence based on prosodic cues such as tone, pace, hesitation, and delivery style.

\vspace{4pt}
{\color{blue!60}\textbf{Classification}}\\
$\bullet$ \texttt{confident}: Clear, decisive, and stable vocal delivery.\\
$\bullet$ \texttt{unconfident}: Shows vocal uncertainty, hesitation, or wavering in delivery.

\vspace{4pt}
{\color{blue!60}\textbf{Output Format}}\\
Classification: \texttt{<confident | unconfident>}

\vspace{4pt}
Now classify the speaker's vocal confidence in the provided audio.
\end{tcolorbox}
\caption{Prompt for vocal confidence classification (audio-only).}
\label{tab:prompt-audio}
\end{table}


\begin{table}[htbp]
\centering
\begin{tcolorbox}[
  width=0.95\columnwidth,
  colback=gray!5,
  colframe=gray!50,
  boxrule=0.6pt,
  arc=3pt,
  left=6pt, right=6pt, top=4pt, bottom=5pt,
  title=\textbf{Few-Shot Speaker Calibration},
  coltitle=white,
  colbacktitle=gray!75,
  fonttitle=\footnotesize,
  fontupper=\footnotesize
]
You are a financial analyst evaluating vocal confidence in earnings call Q\&A.

\vspace{4pt}
{\color{blue!60}\textbf{Input}}\\
You will receive \texttt{\{baseline\_count\}} same-speaker reference Answer audios from the same call, followed by a target Answer audio. Use the reference audio as context when judging the target's vocal confidence.

\vspace{4pt}
{\color{blue!60}\textbf{Classification}}\\
$\bullet$ \texttt{confident}: Shows a clear and stable delivery, with no notable signs of increased uncertainty relative to the reference audio.\\
$\bullet$ \texttt{unconfident}: Shows increased uncertainty relative to the reference audio, such as greater hesitation, shakiness, or instability in delivery.

\vspace{4pt}
{\color{blue!60}\textbf{Output Format}}\\
Classification: \texttt{<confident | unconfident>}\\
Reasoning: \texttt{<single sentence describing the vocal cues supporting the decision>}

\vspace{4pt}
Reference Answer audios:\\
Target Answer audio:
\end{tcolorbox}
\caption{Prompt for few-shot vocal confidence classification with same-speaker reference audio.}
\label{tab:prompt-fewshot}
\end{table}


\begin{table}[htbp]
\centering
\begin{tcolorbox}[
  width=0.95\columnwidth,
  colback=gray!5,
  colframe=gray!50,
  boxrule=0.6pt,
  arc=3pt,
  left=6pt, right=6pt, top=4pt, bottom=5pt,
  title=\textbf{Speaker-Normalized Vocal Confidence Classification},
  coltitle=white,
  colbacktitle=gray!75,
  fonttitle=\footnotesize,
  fontupper=\footnotesize
]
You are a financial analyst evaluating vocal confidence in earnings call Q\&A.

\vspace{4pt}
{\color{blue!60}\textbf{Input}}\\
You will receive three same-speaker reference Answer audios from the same call, followed by one target Answer audio.

\vspace{2pt}
\textbf{Step 1.} Infer the speaker's typical delivery from the reference audios, considering speech rate, pitch/volume stability, articulation, prosody, fillers/pauses, and cadence.\\
\textbf{Step 2.} Judge whether the target shows reduced vocal confidence relative to this speaker-specific baseline.

\vspace{4pt}
{\color{blue!60}\textbf{Classification}}\\
$\bullet$ \texttt{confident}: The target remains broadly consistent with the speaker's typical delivery, without notable increases in hesitation or instability.\\
$\bullet$ \texttt{unconfident}: The target shows increased uncertainty relative to the speaker's baseline, reflected in changes such as hesitation, shakiness, unstable prosody, or disrupted delivery.

\vspace{4pt}
{\color{blue!60}\textbf{Output Format}}\\
Classification: \texttt{<confident | unconfident>}\\
Reasoning: \texttt{<single sentence identifying the main vocal cues and comparing the target to the speaker's baseline>}

\vspace{4pt}
Reference Answer audios:\\
Target Answer audio:
\end{tcolorbox}
\caption{Prompt for speaker-normalized vocal confidence classification with explicit speaker-baseline inference.}
\label{tab:prompt-speaker-norm}
\end{table}

\end{document}